# Point Cloud Registration for LiDAR and Photogrammetric Data: a Critical Synthesis and Performance Analysis on Classic and Deep Learning Algorithms

*Ningli Xu [a,c], Rongjun Qin [a,b,c,d,\*], and Shuang Song [a,b]*

[a] *Geospatial Data Analytics Lab, The Ohio State University, Columbus, USA;*

[b] *Department of Civil, Environmental and Geodetic Engineering, The Ohio State University, Columbus, USA;*

[c] *Department of Electrical and Computer Engineering, The Ohio State University, Columbus, USA;*

[d] *Translational Data Analytics Institute, The Ohio State University, Columbus, USA;*

*\*Corresponding author:* qin.324@osu.edu

**Abstract**

Three-dimensional (3D) point cloud registration is a fundamental step for many 3D modeling and mapping applications. Existing approaches are highly disparate in the data source, scene complexity, and application, therefore the current practices in various point cloud registration tasks are still ad-hoc processes. Recent advances in computer vision and deep learning have shown promising performance in estimating rigid/similarity transformation between unregistered point clouds of complex objects and scenes. However, their performances are mostly evaluated using a limited number of datasets from a single sensor (e.g. Kinect or RealSense cameras), lacking a comprehensive overview of their applicability in photogrammetric 3D mapping scenarios. In this work, we provide a comprehensive review of the state-of-the-art (SOTA) point cloud registration methods, where we analyze and evaluate these methods using a diverse set of point cloud data from indoor to satellite sources. The quantitative analysis allows for exploring the strengths, applicability, challenges, and future trends of these methods. In contrast to existing analysis works that introduce point cloud registration as a holistic process, our experimental analysis is based on its inherent two-step process to better comprehend these approaches including feature/keypoint-based initial coarse registration and dense fine registration through cloud-to-cloud (C2C) optimization. More than ten methods, including classic hand-crafted, deep-learning-based feature correspondence, and robust C2C methods were tested. We observed that the success rate of most of the algorithms are fewer than 40% over the datasets we tested and there are still are large margin of improvement upon existing algorithms concerning 3D sparse corresopondence search, and the ability to register point clouds with complex geometry and occlusions. With the evaluated statistics on three datasets, we conclude the best-performing methods for each step and provide our recommendations, and outlook future efforts. The evaluated dataset and methods are available at
https://github.com/GDAOSU/Awesome-PointCloudRegistration

## 1. Introduction

Point cloud registration remains a fundamental task in the fields of photogrammetry, computer vision, and robotics. Given two sets of point clouds in arbitrary coordinate systems, the goal is to estimate geometric transformations between them to precisely align these data under the same coordinate frame (Tam et al., 2013). This seemingly straightforward task often becomes intractable as the data volume, quality, scale, and scene complexity vary. In geomatics and photogrammetry applications, the sources of data can range from millimeter-level accuracy mobile laser scanning point clouds, indoor scanning data, to half-meter resolution satellite photogrammetric point clouds, with drastically different uncertainty associated with these point measurements. Typical point cloud registration works focus on registering same-source and cooperative datasets, such as registration between two terrestrial laser scanning (TLS) point clouds, or between airborne laser scanning (ALS) point clouds, etc. In these applications, scans are well prepared to have sufficient overlap, metric accuracy, similar density and resolution (scale) for registration and stitching, thus the standard registration algorithms, such as the iterative closest point (ICP) (Arun et al., 1987; Besl & McKay, 1992) or least squares three-dimensional (3D) surface matching methods (Rusu et al., 2009), can easily succeed. However, beyond these standard registration applications, many applications require registering non-cooperative datasets, such as different-source, low-overlap, or metrically inaccurate datasets. For example, in 3D data assimilation and spatial referencing, it is essential to co-register multi-source datasets, such as registration between TLS scans to drone photogrammetry point clouds, between drone scans with satellite photogrammetry point clouds, and between indoor scans with outdoor scans (Djahel et al., 2021). These tasks are extremely challenging and can easily fail standard registration algorithms. Fig. 1 shows several examples of point cloud registration tasks containing both cooperative (Fig. 1 (a-b)) and non-cooperative cases (Fig. 1 (c-e)): 1) registration between two ALS scans from the same source, which shares a large overlap (Fig. 1 (a)); 2) registration between two outdoor TLS scans with sufficient overlaps (Fig. 1 (b));  3) registration between drone and satellite point clouds with different resolution and accuracy (Fig. 1 (c)): 4) registration between drone and satellite point clouds, where the drone point clouds are subject to topographical distortions due to incorrect camera geometry and lack of ground control (Fig. 1 (d)); 5) registration between a mobile laser scanning (MLS) point cloud and a simultaneous-localization-and-mapping-based (SLAM-based) point cloud to significant trajectory drift due to no loop closure.

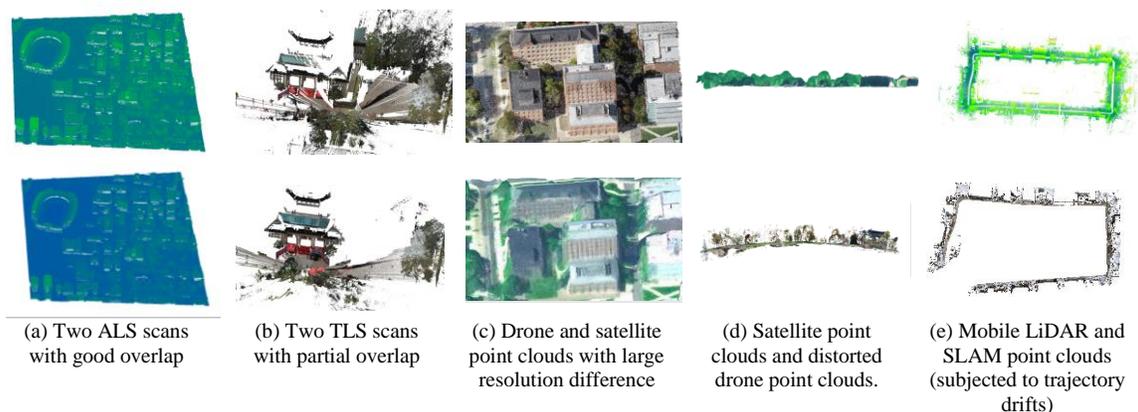

(a) Two ALS scans with good overlap   (b) Two TLS scans with partial overlap   (c) Drone and satellite point clouds with large resolution difference   (d) Satellite point clouds and distorted drone point clouds.   (e) Mobile LiDAR and SLAM point clouds (subjected to trajectory drifts)

**Fig. 1.** Examples of point cloud registration scenarios with increasing difficulties. (a-b) show two examples of cooperative datasets and (c-e) show non-cooperative datasets. Detailed descriptions of this fig.  are in the texts.

Despite the many studies on point cloud registration (Ao et al., 2021; Cai et al., 2019; Yang et al., 2021), due to the highly disparate registration tasks with their respective challenges (as shown in Fig. 1), these methods are mostly tailored to be task specific. Therefore, prior surveys either focus on reviewing methods of point clouds registration on a specific type of data (e.g., TLS data (Dong et al., 2020), multi-angle light detection and ranging (LiDAR) data (Cheng et al., 2018)), or certain technical components of a general point cloud registration pipeline such as 3D point feature extraction and outlier filtering for 3D matching, lacking a full lens of the topic on geomatics and photogrammetry related applications. Thus, from a practical perspective, it is still unclear how these methods perform under varying application scenarios (such as those depicted in Fig. 1). For example, a recent review (Dong et al., 2020) focused on a literature synthesis on the point clouds registration of TLS datasets, where they provided a general technical overview of existing registration methods, where theoretical analysis was provided and a new TLS benchmark dataset for registration was released. However, no comparative experiments were conducted for performance evaluation on this dataset.

In general, most of the point cloud registration pipelines inherently assume a two-step process, 1) a coarse registration establishing an initial geometric transformation between the two datasets using sparse feature-based correspondences, and 2) a fine registration to refine the geometric transformation using more and denser correspondences, also known as cloud-to-cloud (C2C) registration. Depending on the application scenarios, sometimes either of the two steps may be given a higher emphasis, for example, coarse registration is not necessarily a challenge when registering data with relatively accurate initial positions either from geo-referenced data using global positioning system (GPS) measurements or from manual measurements, thus the focus is more on the fine registration; when registering non-georeferenced point clouds, the coarse registration, i.e., identifying correct 3D correspondences for estimation geometric transformation is then the key to success (Dong et al., 2020).

To comprehend the performances of various registration algorithms, there has been a line of work focusing on evaluating their accuracy and robustness. However, most of these evaluations either focus on evaluating certain technical components out of the full registration solution (Guo et al., 2016; Kim & Hilton, 2013; Restrepo & Mundy, 2012; Tombari et al., 2013; Yang et al., 2021) or using limited scale datasets (e.g., synthetically-generated data, structure-light or LiDAR scans of indoor objects) (Guo et al., 2016; Pomerleau et al., 2013; Rusinkiewicz & Levoy, 2001). These evaluation studies include performance analysis on keypoint detection & description algorithms (Guo et al., 2016; Tombari et al., 2013), and outlier-removal algorithms for transformation estimations (Yang et al., 2021). In conclusion, they identified the best-performed algorithms and registration strategies (accuracy-wise and speed-wise), often out of a dataset coming from a specific source. These evaluations provided key insights into these technical components, yet it is understood that a successful registration application requires the robust performance of all these technical components on sufficiently complex and diverse datasets. Moreover, many of these early evaluations do not consider recently developed methods, especially those developed under deep learning paradigms.

In this paper, we aim to close these gaps by providing a comprehensive review of the existing point cloud registration methods containing both the feature-based coarse registration and the C2C fine registration methods, coupled with their performance on three diverse datasets consisting of both same-source and different-source point clouds. The three datasets contain two publicly available TLS datasets containing hundreds of cooperative and non-cooperative examples, and one dataset consisting of point clouds from three sources, drone, satellite photogrammetry, and airborne LiDAR. As compared to existing review and performance evaluation, we draw a comprehensive picture of the technical landscape of the point cloud registration methods (including the most recent deep-learning-based methods), as well as their accuracy and robustness as the whole testing against hundreds of point clouds pairs varying in scale, overlaps and density. To our best knowledge, this is the first work that surveys and evaluates deep learning 3D registration algorithms on photogrammetric and LiDAR point clouds. Our main contributions can be summarized are as follow:

- We provide a comprehensive review of the existing point cloud registration methods with a focus on approaches for practical photogrammetric applications.

- We evaluate the state-of-the-art (SOTA) registration methods using three challenging datasets with various 3D mapping tasks including registration of LiDAR and photogrammetric point clouds that are similar or drastically different in scales, registration of multi-source point clouds such as matching TLS scans with satellite/aerial-based point clouds.
- We investigate the most recent deep learning registration algorithms (as of 2022) and their generalization capability on photogrammetric and LiDAR point clouds.

The rest of this paper is organized as follows: in Section 2 we provide an overview of the existing registration methods; in Section 3 we introduce the evaluation methodology by describing the selected methods, datasets, challenges, and metrics; in Section 4, we provide the experimental results and corresponding analysis; finally, in Section 5 we summarize, discuss the main findings, and conclude our work.

## 2. An overview of the 3D point cloud registration methods

As mentioned above, a typical point cloud registration application inherently assumes two registration steps: 1) a feature-based coarse registration and 2) a C2C fine registration. These two steps are defined as follows: 1) feature-based registration: 3D sparse feature correspondence search followed by robust estimators for resolving a rigid or similarity transformation, often used for coarse registration. The key of this step is to generate highly accurate and robust 3D correspondences between two datasets; 2) cloud-to-cloud fine registration: assuming an initial registration of point clouds, the C2C fine registration starts with an identity transformation to yield pointwise correspondences and iteratively refines the transformation such that the C2C distances are minimized. This type of method is mostly represented by the well-known ICP algorithms (Arun et al., 1987; Besl & McKay, 1992). Some of the methods model the stochastic distance between point clouds, these are regarded as probabilistic methods (Jian & Vemuri, 2011; Myronenko & Song, 2010). The technical landscape of the point cloud registration methods rooted in these two steps is shown in Fig. 2. These methods have further branches detailing typical methods associated with them, which are introduced in the following subsections.

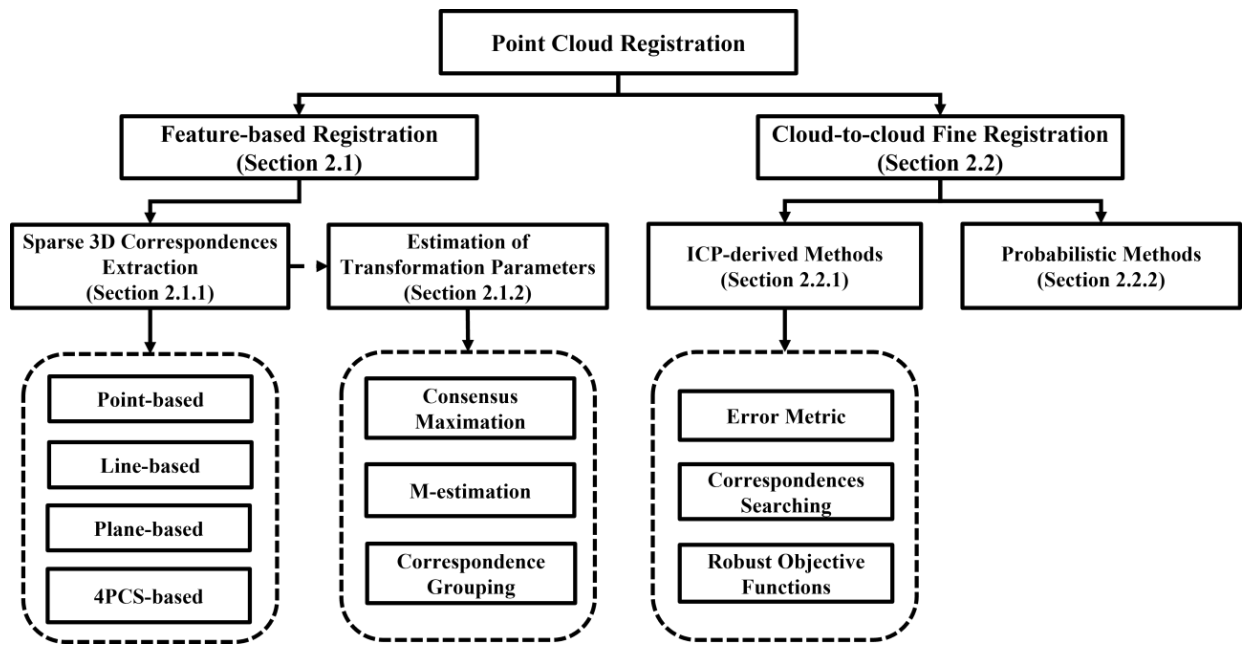

**Fig. 2.** An overview of the point cloud registration methods covered in this paper. Details of these methods are in the texts.

## 2.1. Feature-based registration methods

Feature-based registration methods aim to first find 3D correspondences between two datasets and then estimate the geometric transformations between these correspondences. In general, these 3D correspondences are assumed to be a pair of 3D locations with distinctive features on the object surface, and their presences are usually regarded as "sparse". Since the estimated transformation is fully dependent on these 3D correspondences, ideally, they are expected to be outlier free. Therefore, feature point extraction and matching are critical to feature-based registration methods. Generally, distinctive features in a 3D dataset are extracted based on geometric primitives such as key points, lines, and planes. However, matching features between two datasets can be extremely challenging and usually results in a large number of outliers, mainly due to that these geometric features not distinctive nor robust enough. In practice, the correct correspondences only account for a small portion out of the whole set, in extreme cases, they can be as low as 5% (Bustos & Chin, 2018). Thus, a good registration method must be robust to tolerate a large portion of outliers. There have been efforts tackling these challenges, and we will introduce the efforts in sparse 3D correspondences extraction (Section 2.1.1) and robust estimator for geometric transformations (Section 2.1.2) to constitute a successful feature-based registration.

### 2.1.1. Sparse 3D correspondences extraction

Feature extraction can be achieved by primitive detection methods that contain different types of geometric features, such as key points (Bai et al., 2020; Huang et al., 2021; Li & Lee, 2019; Sipiran & Bustos, 2011; Zhong, 2009), lines (Chen & Yu, 2019; Guru et al., 2004; Tao et al., 2020), and planes (Deschaud & Goulette, 2010; Fotsing et al., 2021; Yang et al., 2010).

Key points are the most widely used 3D features due to their simplicity, whereas for urban scenes and man-made objects, line-based or plane-based representation can achieve better performance (Chen et al., 2020). To match these extracted geometric features, a process called feature description is used to encode unique information about the features (key points, lines, or planes), which is usually done by extracting local characteristics from the local surface reflecting information related to 3D location, gradients and appearance (if color channels are applicable) (Ao et al., 2021; Brenner et al., 2008; S. Chen et al., 2020; X. Chen & Yu, 2019; Faugeras & Faugeras, 1993; Guo et al., 2013; Huang et al., 2021; J. Li, Zhan, et al., 2022; Rusu et al., 2009; Stamos & Leordeanu, 2003; Tao et al., 2020). This process has been advanced by data-driven methods replacing hand-crafted feature descriptors with learning-based descriptors. Most of the works are based on point-based features due to the simplicity of point primitive (Ao et al., 2021; Bai et al., 2020; Huang et al., 2021). Once features and their descriptions are extracted, correspondences can be found by finding the most similar features between two datasets. The similarity metric is then defined in a vector space (e.g., hamming distance, Euclidean distance). To reduce the outliers, this matching process can be further constrained by augmenting the distinctiveness of the correspondence through standard methods such as ratio test (Lowe, 2004), spectral matching (Leordeanu & Hebert, 2005), or tuple test (Zhou et al., 2016).

**Point-based methods** are the most popular due to their simplicity. Among many of the existing works, fast point feature histograms (FPFH) (Rusu et al., 2009) is a widely used hand-crafted (classic) descriptor that leverages both speed and robustness for complex scene contexts (Kim & Hilton, 2013; Restrepo & Mundy, 2012). The method first computes a simplified point feature histogram (SPFH) for each point and its neighboring points within the local surface, and then these SPFH are linearly combined to construct a 33-dimensional feature vector. More recent methods mostly focused on extracting deep features to construct descriptors. SpinNet (Ao et al., 2021) is a learning-based point feature descriptor, which firstly transforms the local surface patch into a rotation-invariant cylindrical volume followed by a 3D cylindrical convolution network (3DCCN) to extract descriptions. It was reported as the top performer in a few indoor and outdoor benchmark datasets at its time of publication (Geiger et al., 2012; Zeng et al., 2017). Geometric transformer

(Qin et al., 2022) learns the robust geometric feature for matching down-sampled super points. It matches the super points by measuring the overlap of their local neighborhood, and then these matches are propagated into the individual points to construct dense point correspondences. Their experiment showed it was robust for data with low overlap cases.

**Line and plane-based methods** utilize line/curve/plane features which are notably more distinctive since they depict more complex geometry. In general, these features are built upon pairs of point features. Chen & Yu (2019) proposed a 3D curve line detection method for LiDAR point clouds. It first detects the feature points located on the boundary line and intersection line (called fold line) of two planes as key points. These key points are then connected to their local neighborhoods to extract geometrical features such as deflection angles and point distances. These points are then ordered by this information to form conformed point sets and fitted through the improved cubic b-spline curving fitting algorithm. Some methods directly utilize extracted 3D lines and planes as the features. For example, Chen et al (2020) proposed a robust plane/line descriptor assuming that data depicting an area with man-made structures. The method first applies well-practiced line/plane detection methods (Lin et al., 2017; Schnabel et al., 2007) to detect features, and then the descriptors of these features are constructed based on two geometric primitives, i.e., distances and angles between intersecting lines and planes of the quadruplet of intersecting planes, which concludes 8-dimensional feature vectors. It was reported to have notably improved the 3D feature matching correctness.

**Four Point Congruent Set (4PCS)-based methods** are another popular family of methods that match point clouds that follow a certain topological relationship instead of performing the pointwise correspondence search. It aims to extract co-planar points in a group of four as the basic unit for point correspondence matching. The underlying assumption of this approach is that a group of co-planar points under certain topological constraints have a higher likelihood to resist outliers. The 4PCS defines two sets of points (four each) to be congruent under a rigid transformation if their cross ratios are similar (Aiger et al., 2008). This led to a series of registration algorithms utilizing this concept (Aiger et al., 2008; Li et al., 2021; Mellado et al., 2014; Theiler et al., 2014). The limitation of the original algorithm (Aiger et al., 2008) is its quadratic time complexity to extract all 4PCS, which makes it difficult for large-scale datasets. Super4PCS (Mellado et al., 2014) reduced such time cost to linear complexity by using a smart indexing data structure, which is constructed by effectively finding all point pairs locally given distance range. K-PCS, Semantic K-PCS, and Super Edge 4PCS (Ge, 2017; Li et al., 2021; Theiler et al., 2014) utilized the concept of point sets, which grouped individual points as the basic unit for registration, and reported that this leveraged well between the time and accuracy. These algorithms vary with approaches of how these point sets are grouped and pre-processed, for example, K-PCS extracts the stable key points with the 3D Difference-of-Gaussians (DoG) detector (Theiler et al., 2014); Semantic K-PCS assumes classified point clouds, and it first extracts non-ground points, then gradually elevates the layer of detection in the vertical direction to detect more points, finally, it connects and groups these points by their semantic information (Ge, 2017). Super Edge 4PCS assumes certain overlaps between the input data, and extracts and constructs 4PCS only in the overlapped region to reduce the time complexity (Li et al., 2021).

*2.1.2. Estimation of transformation parameters from 3D correspondences*

Estimating a 3D rigid transformation between two point clouds requires as few as three pointwise correspondences (Arun et al., 1987), while sparse 3D correspondences through feature extractions often contain a large number of outliers. The outlier rates of 3D correspondences are extremely high (e.g., as high as 95%), thus typical estimators such as the standard M-estimators (De Menezes et al., 2021) or random sampling consensus (RANSAC) algorithms (Raguram et al., 2008) require re-adaptations. In general, there are three types of robust estimators/frameworks: 1) consensus maximization (including the RANSAC family), 2) M-estimator, and 3) correspondence grouping.

***Consensus maximization*** is a more generalized framework of approach with stochastic nature. It tests multiple hypotheses in the solution space and uses each hypothesis to yield some inliers that comply with this transformation. A set of inliers respecting a hypothesis is called a consensus set, and the solution is found as the hypothesis with the maximal consensus set. This is an NP-hard problem (Chin et al., 2018; Tzoumas et al., 2019), thus researchers seek approximate solutions. The chief one is the standard RANSAC and its variants (Barath & Matas, 2018; Chum et al., 2003; Fischler & Bolles, 1981). Instead of testing every single hypothesis in the solution space, the RANSAC family starts by randomly testing hypotheses and computes a consensus set for each test, and selects the hypothesis with the largest consensus set. It has been proven that it is effective in dealing with correspondences of outlier ratios of less than 80% (Yang et al., 2021). However, when dealing with extremely noisy correspondences of high outlier ratios (greater than 90%), the exponentially growing number of iterations significantly increases the computing cost, while the stochastic nature of this approach may yield unreliable results with fewer iterations on noisy datasets. Therefore, removing outliers for the consensus set is crucial. As one of the RANSAC variants, Graph-Cut (GC) RANSAC (Barath & Matas, 2018) estimates more geometrically accurate parameters than the standard RANSAC with the proposed graph-cut process (Zabih & Kolmogorov, 2004). One advantage of this method is that it further corrects wrongly estimated inliers of the consensus set, which constrain spatial coherences of points using graph cuts. To identify the correct and wrong inliers of the consensus set, graph-cut is used based on their spatial coherence. The 4PCS-based methods (Aiger et al., 2008; Li et al., 2021; Mellado et al., 2014; Theiler et al., 2014) follow a similar process as consensus maximization: they randomly sample the 4PCS correspondences and estimate the rigid transformation for each 4PCS correspondence. The derived solution that registers the largest number of points of the source data to the reference data is considered the final solution. Outliers can also be rejected adaptively by comparing the number of inliers of a hypothesis with the number of inliers of a hypothesis generated using all the observations. As an example, the work of guaranteed outlier removal (GORE) (Bustos & Chin, 2018), ROBIN (Shi et al., 2021) removes the hypothesis that has a smaller number of inliers than that generated using all observations. When the solution space is small (small datasets), the NP-hard problem can be also approached by fast-search methods, such as by enumeration (Enqvist et al., 2012; Olsson et al., 2008), branch-and-bound methods (BnB) (Cai et al., 2019; Campbell et al., 2017; Hartley & Kahl, 2009; Li, 2009; Zheng et al., 2011), tree-search methods (Chin & Koltun, 2019; Chin et al., 2015).

***M-estimator*** is a class of methods that formulate robust cost functions with respect to noisy observations. It includes all the observations (i.e., 3D correspondences) into the adaptive cost function based on fitting residuals. Examples of such works include the Geman-McClure function (Zhou et al., 2016), truncated least squares function (Enqvist et al., 2012; Yang & Carlone, 2019), Huber function (Carlone & Calafiore, 2018), $\ell_1$-norm function (Carlone & Calafiore, 2018)), all of which may require specific solvers such as graduated non-convexity (GNC) (Yang et al., 2020), semidefinite programming relaxation (Blekherman et al., 2012), etc. For example, GNC is used to adaptively optimize its Geman-McClure-based objective function (Zhou et al., 2016). Its optimization process starts with equal-weight observations, then gradually reduces the weights of observations with large residuals (or outliers). It was reported as robust when the outlier rate is less than 80%. GNC was also used to optimize the objective functions based on truncated least squares (Yang et al., 2021), and was shown to be more robust (for the outlier rate of less than 95%).

***Correspondence grouping*** methods group 3D correspondences that comply with rigid transformations (usually 6DoF), and these correspondences will be used to derive the final transformation parameters. Among the classic approaches (Buch et al., 2014; Chen & Bhanu, 2007; Leordeanu & Hebert, 2005; Yang et al., 2019), a comprehensive evaluation (Yang et al., 2021) concluded that the approach of (Chen & Bhanu, 2007) is among the best, while it is found to be sensitive to overlapping ratios. Recent learning-based methods including PointDSC (Bai et al., 2021), deep global registration (DGR) (Choy et al., 2020), and deep hough voting (DHV) (Lee et al., 2021) provided a certain remedy to this drawback, as it was proven that these learnable features encode contextual information to be more accurately and robustly describe correspondence potentials before registration (Bai et al., 2021; Choy et al., 2020). For example, PointDSC (Bai et al., 2021)

uses a few neural network-based modules to map and group initial correspondences at the deep feature level. Correspondences from the same group are assumed to share similar geometric consistency and are tested for registration, and the solution agreeing with most correspondences is selected as the estimated solution.

### 2.2. Cloud-to-cloud fine registration methods

Assuming an initial registration is available, the C2C fine registration methods iteratively update the transformation parameters by optimizing the C2C distance based on dense correspondences between two point clouds. The C2C distance is calculated at the per-point level based on the updated transformation parameters. Since it considers all points in the point clouds, the C2C methods are usually regarded as a final step of registration pipelines to provide the most accurate result. Based on the use of C2C distance, we can divide existing methods into two categories: 1) ICP-derived methods; 2) Probabilistic methods. The ICP-derived methods have been popular in the literature due to their simplicity and effectiveness, while their objective function depends on initial registration and the complexity of point clouds, thus its convergence is not guaranteed. The probabilistic methods model the raw point clouds as samples from certain probability distributions (e.g. gaussian mixture model (GMM) (Jian & Vemuri, 2011; Myronenko & Song, 2010)). The C2C distance can be either the likelihood between observed data and the assumed statistical model or the distance between two statistical models. This effectively serves as a regularization process that will further improve the robustness of the registration over typical ICP-derived methods (Gao & Tedrake, 2019; Liu et al., 2021), and the drawbacks of those approaches are their high-memory consumption and computational complexity, limiting their usage in large-scale datasets.

#### 2.2.1. ICP-derived methods

The standard ICP uses chamfer distance (Barrow et al., 1977) as the C2C distance, which computes the sum of distances of a point in one dataset to its nearest point in the other. The chamfer distance can be easily established by measuring the distance from one point of a cloud to its nearest point of another cloud subject to initial registration. ICP uses the expectation maximization (EM) algorithm (Dempster et al., 1977) to minimize the sum of the squared distances: in each iteration, it establishes the chamfer distance for each point with the current registration parameters, then estimates the registration parameters with respect to the updated chamfer distance. The terminal criterion is the change of C2C distance between two neighboring iterations below a certain threshold. The variants of the ICP algorithm mostly target their improvement on the iterative steps, including improving the robustness, convergence speed, and computation efficiency (Rusinkiewicz, 2019; Zhang et al., 2021). Specifically, these variants focus on improving upon three aspects, 1) better error metric (or loss function), 2) enhanced correspondences searching strategy, and 3) more robust objective functions.

**Improving error metrics:** Point-to-plane ICP (Chen & Medioni, 1992) incorporates surface normal into the error metrics, like $(p_i - q_i) \cdot n_{qi}$, where $(p_i, q_i)$ is a pointwise correspondence, $n_{qi}$ is the surface normal of $q_i$. For each correspondence, the estimated registration tries to align point $p_i$ from one cloud to the plane defined by the point $q_i$ from another cloud. It was reported more accurate with faster convergence speed than the standard ICP (Besl & McKay, 1992). Symmetric ICP (Rusinkiewicz, 2019) proposes a symmetric error metric that incorporates surface normals of both points from pointwise correspondence, like $(p_i - q_i)(n_{pi} + n_{qi})$ to increase the robustness. Generalized ICP (Segal et al., 2009) unifies the standard ICP and point-to-plane ICP variant into a single probabilistic framework and can be considered a soft plane-to-plane ICP. For each correspondence, it models the local surfaces of both points using covariance matrices.

**Improving correspondences searching strategy**. Correspondences generated by nearest neighboring searching are not always correct and often leads to the local minimal solution. Deng et al (2021) generated the correspondences by connecting points from two point clouds using random straight lines. Consistent lines provide cues of correct correspondence, and as an additional input, incorrect correspondences will not be considered.

**Designing more robust objective functions**. Robust ICP (Zhang et al., 2021) added a robust kernel, Welsch's function (Equation(1)), to give less weight to the outliers through a variable bandwidth factor ν. ν starts with a large value $ν_{max}$, halves to a targeted $ν_{min}$, to gradually eliminate outliers at different magnitude.

$$Welsch's\ Function: \rho(r) = 1 - \exp\left(-\frac{r^2}{2\nu^2}\right) \qquad (1)$$

More recent variants of ICP focus on making holistic improvements leveraging accuracy, efficiency, and robustness. For example, robust symmetric ICP (J. Li, Hu, et al., 2022) combined a symmetric error metric with a robust kernel, which showed better robustness than some existing baseline methods (Rusinkiewicz, 2019; Zhang et al., 2021). Moreover, ICP variants were adapted to specific tasks such as odometry applications. For example, KISS-ICP (Vizzo et al., 2022) added adaptive thresholding for correspondence searching, a robust kernel, and a subsampling strategy to standard ICP. Its experiments show demonstrate relatively higher efficiency and robustness in LiDAR odometry applications. Assuming that the point clouds are generated sequentially as the robot moves, it can estimate the pose of the current 3D scan based on the poses of the previous two-time steps. The estimated pose as the initial registration facilitates the standard ICP to find pointwise correspondences and reject outliers.

*2.2.2. Probabilistic methods*

The idea of probabilistic methods is to model the point clouds as certain probability distributions. Thus, the point cloud registration problem is reformulated as a problem to minimize two data distributions. Most existing methods use GMM to model the point clouds, where point clouds can be either the samples of the assumed GMM model or the GMM itself (by taking the location of each point as the centroid). The minimization can be achieved through maximal likelihood estimators (MLE) represented by GMM (Myronenko & Song, 2010). Oftentimes the same GMM is used to model two point clouds, which was reported to be more sensitive than using separate GMMs (Jian & Vemuri, 2011). As compared to ICP-derived methods, probabilistic methods construct more densely weighted correspondences by conceptually assigning more point-to-point connections (as samples), which is more robust yet requires more memory and computing. Therefore, recent works mainly focus on improving their computation efficiency by techniques such as efficient data structures (e.g. GMM-Tree (Eckart et al., 2018)), fast and approximate methods (e.g. fast Gauss transform (FGT) (Myronenko & Song, 2010), permutohedral filter (Gao & Tedrake, 2019)), GPU acceleration (Eckart et al., 2018; Liu et al., 2021). In addition, instead of modeling each point as the centroid of the Gaussian mixture component, recent learning-based methods focus on partitioning the point clouds into several groups of points and model each group as a Gaussian mixture component, such as deep GMM (Yuan et al., 2020), deep hierarchical GMM (Hertz et al., 2020), which showed promising registration performance and speed compared with traditional methods. For example, Coherent point drift (CPD) (Myronenko & Song, 2010) is a widely used probabilistic method for both rigid and non-rigid point cloud registration. It regards the registration of two point clouds as the probability density estimation problem, where one point cloud is assumed a GMM model and the matching point clouds are transformed samples, estimated through MLE. CPD speeds up this procedure with an approximation method so that the computation complexity is reduced to linear time (Greengard & Strain, 1991).

3. **Performance assessment for point cloud registration methods**

In this section, we conduct a comprehensive evaluation of point cloud registration methods using a diverse set of data, including two public datasets and one carefully collected dataset of different sources.

### 3.1. Selected methods for performance evaluation

We select representative algorithms from each category of both feature-based and C2C-based methods. The selection is based on the availability of source codes, as well as their popularity and freshness. As a result, eight feature-based methods and seven C2C-based methods are selected for evaluation, as shown in Table. *1*. The selection leverages both classic and recent deep-learning-based methods. It should be noted that probabilistic and learning-based C2C fine registration methods were not included in this evaluation, since these methods are not yet practical for large-scale datasets (Dong et al., 2020; Eckart et al., 2018).

**Table. 1.** Parameters and implementation of test methods in our evaluation. $r_{feature_{in}}, r_{feature_{out}}$ represent the feature support radius in the meter for indoor and outdoor datasets, $top\_rate = 0.3$ means selection of top 30% correspondences as inliers, $MNI$ is the maximum number of iterations, $w$ means the weight of uniform distribution.

| Category | Methods | Type | Parameters |
|---|---|---|---|
| Feature-based Coarse Registration | **FPFH** (Rusu et al., 2009) | Classic | $r_{feature_{in}} = 0.4$<br>$r_{feature_{out}} = 4.5$ |
| | **SpinNet** (Ao et al., 2021) | Learning-based | $r_{feature_{in}} = 1$<br>$r_{feature_{out}} = 6$ |
| | **SM** (Leordeanu & Hebert, 2005) | Classic | $top\_ratio = 0.3$ |
| | **RANSAC** (Fischler & Bolles, 1981) | Classic | $MNI = 100,000$ |
| | **GC-RANSAC** (Barath & Matas, 2018) | Classic | $MNI = 100,000$ |
| | **FGR** (Zhou et al., 2016) | Classic | $MNI = 100$ |
| | **TEASER++** (H. Yang et al., 2021) | Classic | $MNI = 10000$ |
| | **PointDSC** (Bai et al., 2021) | Learning-based | - |
| Cloud-to-Cloud Fine Registration | **Standard ICP** (Besl & McKay, 1992) | Classic | $MNI = 100$ |
| | **Point-to-plane ICP** (Arun et al., 1987) | Classic | $MNI = 100$ |
| | **Generalized ICP** (Segal et al., 2009) | Classic | $MNI = 100$ |
| | **Symmetric ICP** (Rusinkiewicz, 2019) | Classic | $MNI = 100$ |
| | **Fast ICP, Robust ICP, and Robust point-to-plane ICP** (Zhang et al., 2021) | Classic | $MNI = 100$ |

To ensure a fair comparison, we applied an adaptive parameter-setting strategy, which explores hyper-parameters for each method that achieves peak performance, and these key parameters were listed in Table. *1*, and the readers are couraged to read the original paper to interpret the meaning of these parameters. It should be noted that the hyper-parameter "MNI" refers to the maximal number of iterations, which should be interpreted differently between the feature-based and C2C methods. The "MNI" refers to the iterations used in the RANSAC-like strategy in feature-based methods, thus it varies with different problems, whereas "MNI" in C2C methods refers to that of the cost function minimization, thus is fixed over all the methods.

### 3.2. Experiment datasets

To provide a comprehensive assessment of existing registration methods, the datasets need to encapsulate: 1. ***diverse application scenarios,*** 2. ***diverse data sources,*** and 3. ***varying levels of scene complexity and occlusions***. Therefore, we selected three datasets: two publicly available datasets (Whu-TLS and RESSO) and an additional self-collected dataset (OSUCampus). These datasets cover various scene contexts, are captured by various sensors, and vary in occlusion and clutter. Fig. 3 shows some examples and the details of these three datasets are listed in Table. *2* and introduced in the following paragraphs.

***Whu-TLS*** (Dong et al., 2020) is a large-scale TLS point cloud registration benchmark for diverse outdoor survey applications. The data is captured by multiple laser scanner systems and covers various environmental contexts with different degrees of overlaps. We choose 6 representative scenes including mountain, park, campus, river bank, underground excavation, and tunnel scenarios where the scene contexts cover structured, semi-structured, and unstructured types. The selected scenes enable us to evaluate the performance of selected methods on various survey applications: river surveys, underground asset management, and landslide monitoring.

***RESSO*** (Chen et al., 2020) is a real-world LiDAR scan dataset with a wide range of overlap ratios from 1% to 90% which is aimed at addressing partial overlap LiDAR data registration. The dataset covers both indoor and outdoor scenes and contains two types of lidar scanners: the handheld scanner and the high-range static scanner.

***OSUCampus*** provides multi-source photogrammetry data captured by multiple platforms including satellites, unmanned aerial vehicles (UAV), and airborne platforms. The main challenge is the large difference in accuracy and scale, as shown in

Table. *3*, thus enabling the test of methods in extreme scenarios where data accuracy, scale, point density, occlusion, and sources are completely different. These data are collected on different dates, and there may exist inconsistent objects due to temporal changes (examples shown in the red rectangle region between the satellite and ALS data in Fig. 3.

**Table. 2.** Details of three datasets for evaluation.

| Dataset | Data Source | #Scenes | #Pairs of Scan | #Pts (million) | Density Difference | Point Accuracy | Main Challenge |
|---|---|---|---|---|---|---|---|
| **Whu-TLS** | TLS | 6 | 68 | 922.45 | Yes | 3-4mm | Scene Contexts |
| **RESSO** | TLS | 15 | 172 | 63.04 | No | 1-4mm | Partial Overlap |
| **OSUCampus** | Satellite, UAV, ALS | 2 | 14 | 41.85 | Yes | 0.14-1.5m | Cross-source |

**Table. 3.** Data accuracy of multiple sources in OSUCampus. The data accuracy of satellite and UAV data is from *(Han et al., 2020)*.

| Sources | Data Accuracy |
|---|---|
| Satellite Data | Average: 1.26 m |
| UAV Data | Average: 0.25 m |
| Airborne LiDAR Scanning Data | Average: 0.14 m |

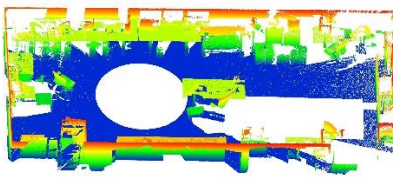
RESSO: indoor-6a

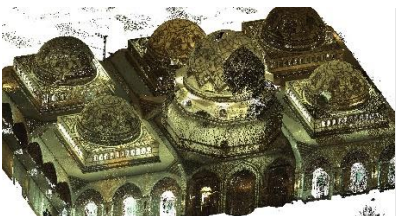
RESSO: indoor-6j

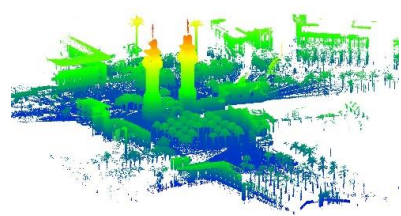
RESSO: outdoor-7a

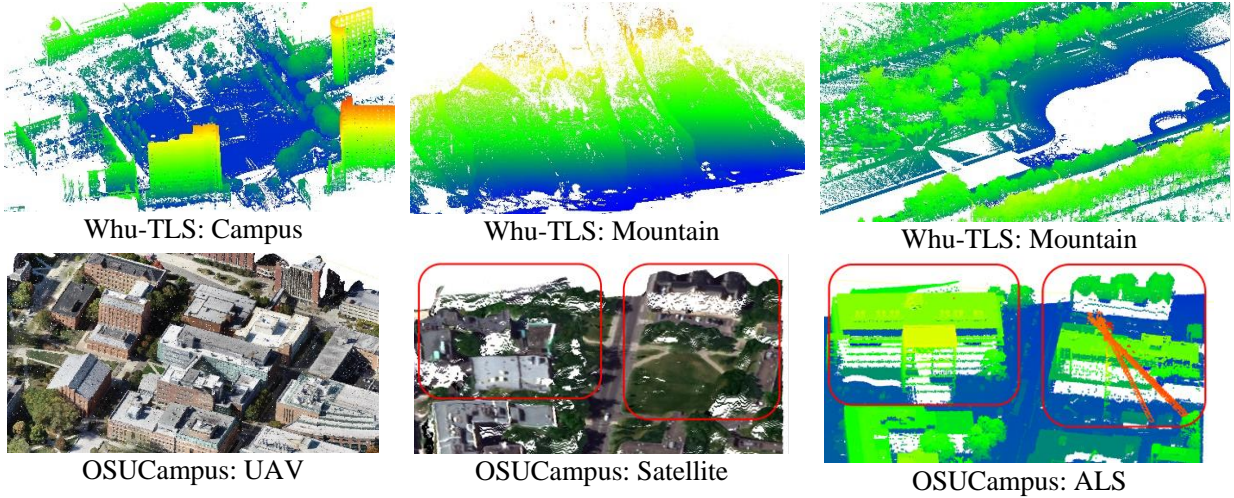

**Fig. 3.** Examples from three datasets. The point clouds without color information are colored according to their height. Volume changes of buildings are marked by red rectangles in OSUCampus: Satellite and ALS.

*3.3. Evaluation metrics*

*3.3.1. Precision-Recall curve (PRC)*

PRC is used to evaluate the descriptiveness of feature descriptors, defined as the precision and recall values under different thresholds (Guo et al., 2016; Ke & Sukthankar, 2004; Tombari et al., 2010). **Precision** and **recall** are two metrics describing the quality of matches, as shown in Equation (2) and Equation (3) (Guo et al., 2016), where matches are generated using the nearest neighbor distance ratio (Lowe, 2004) between features (smaller than 1). For each feature, if the ratio between its nearest and second-nearest neighbors is less than the threshold $\tau_0$, the feature and its nearest neighbor are considered a match. By varying $\tau_0$ from 0 to 1, a series of precision & recall values will be generated and we can plot them into PRC.

$$Precision = \frac{\#\ correct\ matches}{\#\ matches} \quad (2)$$

$$Recall = \frac{\#\ correct\ matches}{\#\ corresponding\ features} \quad (3)$$

*3.3.2. Average root-mean-squared-error (RMSE) and success rate*

We use the RMSE ($err_{rmse}$) between the estimated parameters ($R_{est}, T_{est}$) and ground-truth parameters ($R_{gt}, T_{gt}$) (Equation (4)), as well as the success rate of the registration to evaluate the performance of different algorithms. The success rate measures the percentage of pairs that successfully registered over all pairs. The registration is considered successful if the $err_{rmse}$ is below a threshold $\tau_1$. **Average RMSE** computes the average $err_{rmse}$ overall successfully registered point cloud pairs. The threshold $\tau_1$ may vary with data accuracy, and point density. We follow the suggested values of existing works (Chen et al., 2020; Wang et al., 2021) for the public dataset and adjust appropriately with our dataset: we set $\tau_1$ 0.5m for RESSO indoor data, 1m for RESSO outdoor data, 1.5m for Whu-TLS data, and 2.5m for OSUCampus data.

$$err_{rmse} = \sqrt{\frac{1}{N}\sum_{1}^{N}\left\|R_{est}p_i + T_{est} - R_{gt}p_i - T_{gt}\right\|_2^2} \quad (4)$$

### 3.3.3. $\sigma_{MEAN}$

We use C2C distance $\sigma_{MEAN}$ (Ahmad Fuad et al., 2018) to measure the deviation between the registered point clouds. This is specifically used for our OSUCampus, as no ground-truth registration is available. Specifically, it calculates the average distance of each point to the corresponding local surface as shown in Equation (5). I.e., for each point in the source point cloud, the C2C distance first models the local plane around its nearest six points from the other point cloud and calculates the distance of the point to the plane.

$$\boldsymbol{\sigma_{MEAN}} = Mean(D_{point\_to\_local\_surface}) \quad (5)$$

## 4. Experiment results

In this section, we perform four experiments to evaluate the robustness, accuracy, and efficiency of registration methods (listed in Table. *1*) on three datasets (information shown in Table. *4*). We first evaluate the descriptiveness performance of the selected feature descriptors on three datasets using the PRC. Then we evaluate the robustness and registration accuracy of robust estimators by the success rate and average RMSE. For C2C fine registration methods, we use $\sigma_{MEAN}$ to evaluate their registration accuracy and robustness. Moreover, the time cost of C2C registration methods is also evaluated.

**Table. 4.** Metrics used for each experiment and method category.

| Experiments | Methods | Metrics | Datasets |
|---|---|---|---|
| Descriptiveness (Section 4.1.1) | Descriptors | PRC (Section 3.3.1) | RESSO Whu-TLS OSUCampus |
| Parameter estimation (Section 4.1.2) | Robust estimators | Success Rate Average RMSE (Section 3.3.2) | RESSO Whu-TLS OSUCampus |
| Accuracy (Section 4.2.1) | C2C fine registration | $\sigma_{MEAN}$ (Section 3.3.3) | OSUCampus |
| Computation efficiency (Section 4.2.2) | C2C fine registration | Time cost $t$ | OSUCampus |
| Robustness (Section 4.2.3) | C2C fine registration | $\sigma_{MEAN}$ (Section 3.3.3) | OSUCampus |

### 4.1. Results of feature-based registration methods

### 4.1.1. Performance of feature descriptors

We use PRC (Section 3.3.1) to evaluate the feature descriptors and calculate the area under the PRC curve (AUC) to measure the overall performance of three descriptors, SpinNet (KITTI), SpinNet (3DMatch) and FPFH. It should be noted that SpinNet has two versions trained from different datasets (KITTI (Geiger et al., 2012) and 3DMatch (Zeng et al., 2017)). Fig. 4 shows the results of these algorithms on different datasets. In general, we found that the relative performance of different feature descriptors is consistent throughout different datasets, while some datasets are more challenging than others. Among all the descriptors, SpinNet (KITTI) consistently achieves the best results, while the performances of the other two descriptors vary with the datasets. For example, SpinNet (3DMatch) showed poor results in Whu-TLS dataset, partly due to that 3DMatch is an indoor dataset while the Whu-TLS dataset is mostly outdoor. All descriptors, as expected, performed order-of-magnitude worse in the OSUCampus dataset, due to the extremely different scale and accuracy of different scans.

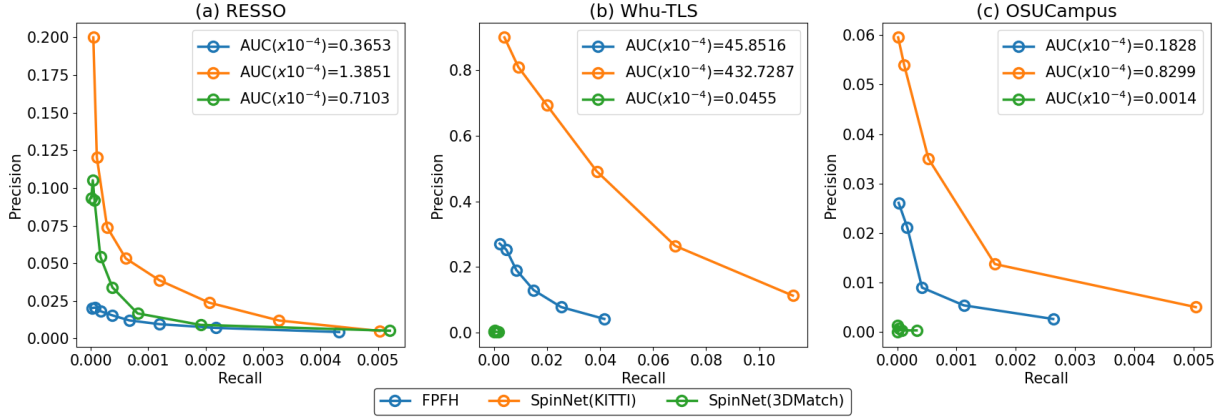

**Fig. 4.** The descriptiveness performance of the selected descriptors on three datasets. SpinNet (KITTI), and SpinNet (3DMatch) represent the SpinNet trained on KITTI Odometry (a street-view LiDAR dataset) and 3DMatch (an indoor RGB-D dataset ) datasets respectively. AUC measures the overall descriptiveness performance of selected descriptors. The bigger AUC is, the more descriptive the test method is on that dataset. Axes are appropriately scaled for visualization.

From the above observations across varying datasets, we can conclude that, first, the learning-based descriptor outperforms the traditional descriptor on three photogrammetric datasets. Among three datasets with different challenges including partial overlap, various scene contexts, and cross sources, the learning-based descriptor, SpinNet trained with the KITTI dataset, constantly outperforms the traditional descriptor, FPFH. Second, the performance of the learning-based descriptor, SpinNet, highly depends on its training datasets. SpinNet (KITTI) achieved the best descriptiveness performance on the three datasets while SpinNet (3DMatch) barely produces correct matches among Whu-TLS and OSUCampus datasets. Third, cross-source data and partial overlap data are two challenging factors for both learning-based and traditional descriptors. Among the three datasets, RESSO and OSUCampus datasets are much more challenging than Whu-TLS, since the AUC results of all descriptors on RESSO (AUC values ($\times 10^{-4}$): 0.3653, 1.3851, and 0.7103) and OSUCampus (AUC values ($\times 10^{-4}$): 0.1828, 0.8299 and 0.0014) datasets are much lower than those on Whu-TLS dataset (AUC values ($\times 10^{-4}$): 45.8516, 432.7287 and 0.0455). These AUC values are very low, due to that sparse feature correspondences and descriptors often have much higher outlier rates.

### 4.1.2. Performance of estimators

We use success rate and average RMSE (Section 3.3.2) to evaluate the six robust estimators introduced in Section 3.1. To make a fair evaluation, we first select the best descriptor in the previous experiment, SpinNet (KITTI), to extract correspondences on three datasets, and then estimate the transformation parameters based on those correspondences using the six robust estimators. It should be noted that similarly to SpinNet, PointDSC has two versions trained from two datasets (KITTI and 3DMatch), denoted as PointDSC (KITTI) and PointDSC (3DMatch). Results on different datasets are shown in Fig. 5. It can be seen that among all three datasets, there is no clear winner, however, we do observe that datasets play a major role in the overall performance of these estimators. For example, most of the estimators consistently achieve good results in Whu-TLS, partly due to that the datasets are mostly well-overlapped and are with accurate scans. The lack of data overlap in RESSO datasets can be the major factor that most of the estimators do not achieve a good success rate (all smaller than 40%). Among these methods, GC-RANSAC achieves the best RMSE. Moreover, the learning-based method PointDSC consistently shows comparatively low RMSE (ranked top-3 in all datasets), indicating that it has a great potential to handle difficult registration cases (i.e., large-scale differences and lack of overlaps). Although the overall success rate is low for RESSO and OSUCampus data,

TEASER ++ constantly achieves relatively the highest success rate results while maintaining moderate accuracy among the three datasets.

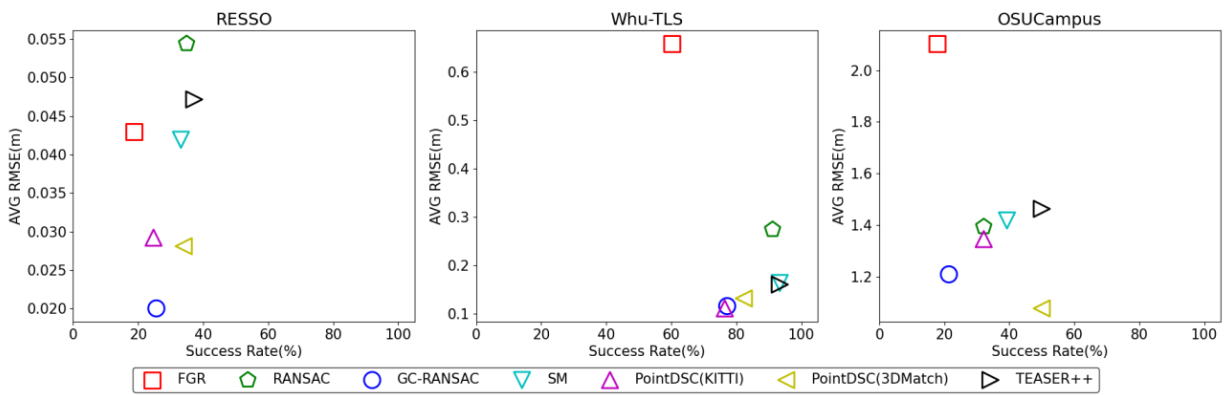

**Fig. 5.** Registration performance of robust estimators on three datasets.

*4.2.    Results of cloud-to-cloud registration methods*

*4.2.1.    Registration accuracy & efficiency performance*

Assuming the same initial registration of test data, we evaluate seven C2C methods in terms of their fine registration performance. Since OSUCampus dataset contains point clouds from diverse sources constructing the most versatile registration pairs, we primarily used this dataset for evaluation. As shown in Fig. 6, this dataset contains UAV, ALS, and satellite-derived (SAT) point clouds at different sizes and scales. Each source data is registered with the rest, yielding three pairs of data in total: UAV-SAT, UAV-ALS, and ALS-SAT. We evaluate the registration accuracy of these methods using $\sigma_{MEAN}$ (details in Section 3.3) as well as their running time.

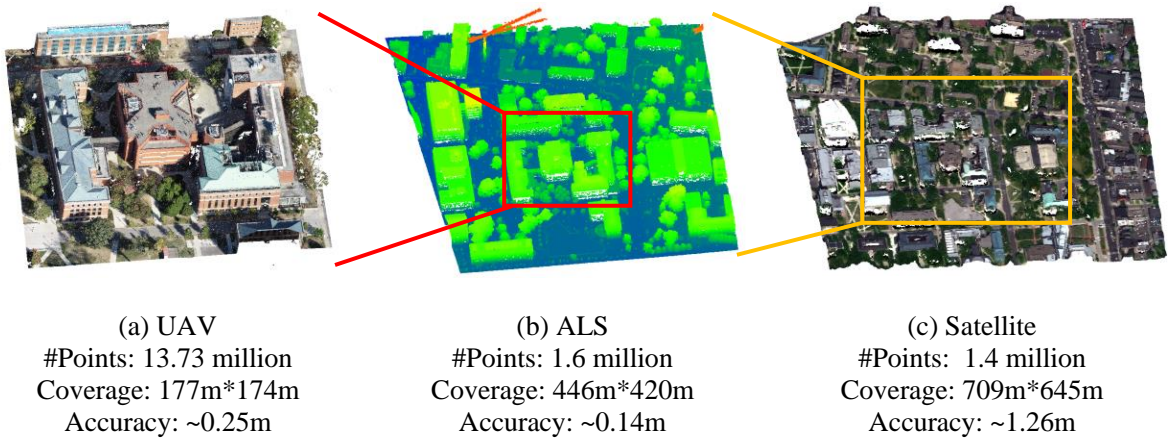

(a) UAV
#Points: 13.73 million
Coverage: 177m*174m
Accuracy: ~0.25m

(b) ALS
#Points: 1.6 million
Coverage: 446m*420m
Accuracy: ~0.14m

(c) Satellite
#Points:  1.4 million
Coverage: 709m*645m
Accuracy: ~1.26m

**Fig. 6.** Test data from three types of sources: UAV, ALS, and satellite.

Table. 5 shows the results of these fine-registration methods. Overall, the Generalized ICP method appears to be a notable approach in terms the achieved accuracy, which leverages both accuracy and speed. It achieves the best accuracy in two out of the three cases while not consuming as much time as others. Symmetric ICP by its results is also a comparably strong approach. It achieves the best accuracy in one of the three datasets, and good results (second best) for the other two datasets, at the same time the fastest convergence time that is ten to hundred times better than the other methods. The point-to-plane ICP seems to fail in all three datasets, given its large $\sigma_{MEAN}$ as compared to others as well as the extremely long computing time (likely due to algorithm divergence). Other approaches, as shown in Table. 5, have their respective advantages, for example, Fast ICP, although not as fast as Symmetric ICP, is relatively efficient over other approaches, while achieving notable results (mostly in top-3).

Table. 5. Registration accuracy and efficiency results of seven test fine-registration methods.

|  | UAV-SAT | | UAV-ALS | | ALS-SAT | |
| --- | --- | --- | --- | --- | --- | --- |
|  | $\sigma_{MEAN}$(m) | t(s) | $\sigma_{MEAN}$(m) | t(s) | $\sigma_{MEAN}$(m) | t(s) |
| **Standard ICP** | 1.9204 | 325.8 | 0.3989 | 155.1 | 2.4849 | 72.7 |
| **Fast ICP** | 1.9203 | 98.7 | 0.3989 | 68.5 | 2.4811 | 7.6 |
| **Robust ICP** | 1.9642 | 2167.8 | 0.3729 | 308.1 | 2.4740 | 186.3 |
| **Point-to-plane ICP** | 5.7038 | 1424.9 | 77.6102 | 2337.4 | 6.8872 | 247.1 |
| **Robust Point-to-plane ICP** | 2.6773 | 773.7 | 1.0094 | 532.2 | 2.8188 | 178.3 |
| **Generalized ICP** | **1.9068** | 152.6 | 0.3721 | 176.8 | **2.4148** | 32.7 |
| **Symmetric ICP** | 1.9173 | **2.5** | **0.3719** | **1.5** | 2.4309 | **0.6** |

*4.2.2. Robustness performance to the initial registration*

C2C fine registration methods are known to be sensitive to the initial registration and bad initial parameters can fail the registration. To evaluate how robustly the C2C methods can perform with respect to different initial registrations, using the OSUCampus dataset, we intentionally rotate and translate the point clouds to create "bad" initializations, and evaluate the performance of different registration algorithms. Specifically, the rotation was performed with respect to the Z axis and the translation was attempted at the X axis. The evaluation results are shown in Fig. 7.

For translation, most of these C2C algorithms can recover translation differences of more than 15 meters for registration on the UAV-ALS and UAV-SAT cases, while most of the algorithms achieved comparable results, Fast ICP performed slightly better and the Generalized ICP slightly worse when dealing with the UAV-ALS case. However, we observed that the performances of these algorithms varied when dealing with the ALS-SAT, where the Robust ICP and Fast ICP outperformed the rest, with the Generalized ICP still the poorest. These large differences in performance can be attributed to the temporal differences between the two sources (ALS and SAT, see Fig. 3, the rectangle regions in the third row). The better performances, in this case, indicate that these algorithms (i.e., Robust ICP and Fast ICP) are better at handling large outliers.

For rotation, the results suggest similar conclusions, but the performance variations of these algorithms showed a slightly different landscape. For example, Robust ICP and Fast ICP still appear to be the best, being able to recover rotations of more than 50 degrees in the best case (UAV-ALS), and consistently outperforming the other algorithms. Generalized ICP is still among the worst, while additionally the Symmetric ICP and standard ICP show similarly poor results. Moreover, the overall performance of these algorithms is poorer in the ALS-SAT case, in which they can mostly tolerate about 20 degrees of differences due to some changes in the content (buildings) in the dataset (see Fig. 3).

It is worth noting that although some of the methods, such as Generalized ICP and Symmetric ICP, achieved the best results (as concluded in Section 4.2.1.) when registration parameters are well initialized, they can be very sensitive to errors of the initial registration parameters. Instead, the moderately performed algorithms in Section 4.2.1, such as Fast ICP and Robust ICP, can tolerate much higher errors in the initial registration parameters, making them a more preferred choice concerning the algorithm robustness.

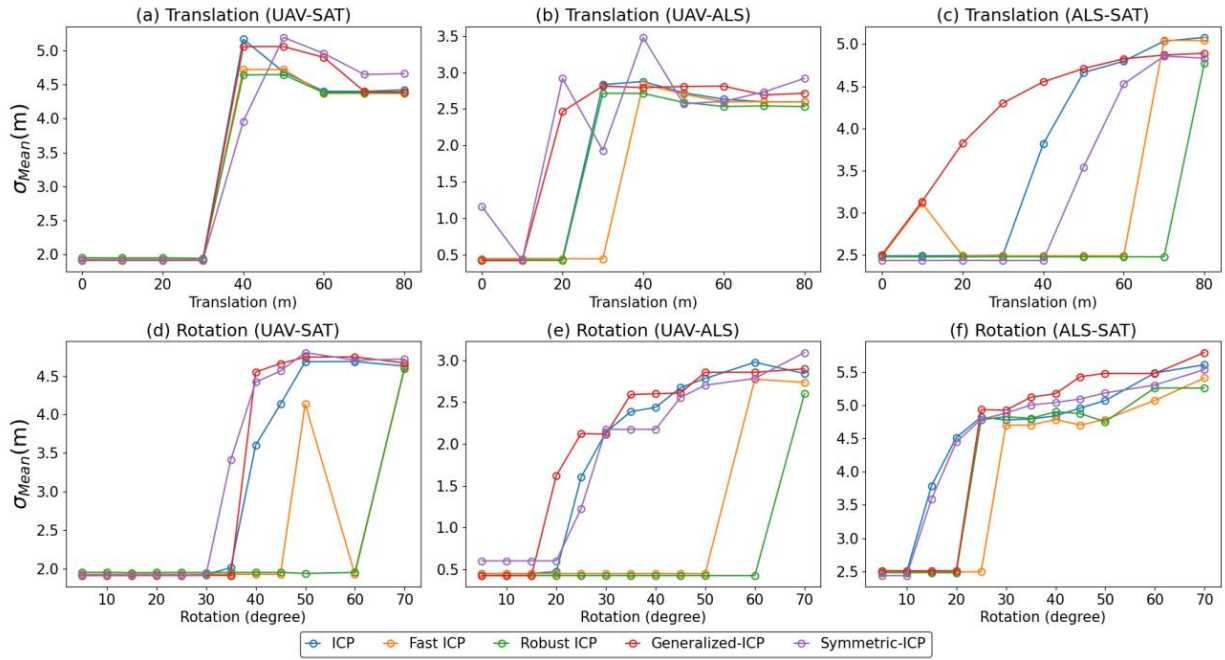

**Fig. 7.** Robustness performance of C2C fine registration methods with respect to different initial registrations. The initial translation is performed along the x-axis while the initial rotation is performed about the z-axis.

## 5. Conclusions and outlook

This paper presents a review and performance analysis of existing point cloud registration algorithms. Specifically, our review considers two separate steps of a registration task, being 1) a coarse registration using sparse feature-based correspondences, and 2) a fine registration through C2C registration. We review both the classic and deep learning-based approaches for sparse correspondence extraction, and matching, as well as prevalent approaches for estimating rigid transformations between registration subject to large numbers of outliers (Section 2).

We have tested these methods on a diverse set of point cloud datasets including two public registration datasets (RESSO and Whu-TLS) and a self-collected dataset (OSUCampus), ranging from indoor to outdoor scans, with sources from airborne LiDAR to satellite photogrammetric point clouds, all datasets are at a varying degree of overlap, scale, and accuracy. Both the coarse registration and C2C-based registration are tested against point clouds from these datasets.

Through intensive experiments (Section 4), we respectively concluded our observations on 1) sparse feature extractors, 2) robust estimators for rigid 3D transformation in support of point cloud registration, and 3) C2C algorithms for fine registration. Based on the experiment results, we concluded that: 1) for coarse registration, deep-learning-based approaches, such as SpinNet, PointDSC, and the classic approach TEASER++ achieved the best results respectively for feature extraction, description, and robust parameter estimation, while we also observed that the achieved accuracy are not sufficient to address many of the point cloud registration challenges, such as low overlap, large scale differences, evidenced by the overall poor performance on two of the challenging datasets (RESSO and OSUCampus). Moreover, learning-based methods may significantly suffer from generalization issues, leading to completely different results when a method (SpinNet) is trained on different datasets. 2) For C2C fine registration, we conclude that Fast ICP and Robust ICP are preferred options given their good leverage between accuracy and robustness. Generalized ICP and Symmetric ICP both achieved the best accuracy, however, are very sensitive to errors in the initial parameters, partly due to their use of surface normals in addition to point distances, which brings another dimension of uncertainty.

**Recommendations.** Through the experiences from the intensive experiment, we provide the following recommendations in practice: for arbitrary datasets with unknown initial registration, we recommend SpinNet to extract the features, then TEASER++, RANSAC, or SM for rough registration estimation, followed by Fast ICP or Robust ICP for fine registration estimation. For datasets with well-estimated initial registration parameters (i.e., less than 10 meters and a few degrees), we recommend the Generalized ICP or Symmetric ICP to fine-tune the registration. For real-time 3D mapping tasks, we recommend FPFH for feature extraction, TEASER++ for rough registration estimation, and Symmetric ICP or Fast ICP for fine registration estimation.

**Outlook.** As mentioned earlier, despite the great advances from the large body of classic and deep learning-based methods, the achievable accuracy and robustness are still insufficient for complex datasets, given the extremely low success rate on challenging datasets. It is worth noting that many of the registration algorithms have consistent performance across varying datasets. While one direction to follow, is to further improve some of the winning methods by adding incremental strategies to outlier removal processes. However, for challenging datasets, existing methods can perform as poorly as 20% of success rate or less, in which the 3D feature extractor and correspondence search are the key. Thus, we consider that deep-learning-based methods can still be further explored and there is a large margin for improvement given the overall poor performance of the existing approaches. C2C-based methods generally performed well given sufficiently accurate initial solutions, while they do suffer the lack of robustness and inability to scale to large-scale datasets. Therefore, future C2C methods should focus on improving the robustness and running time by incorporating more robust optimization algorithms.

## 6. Acknowledgments

The authors are supported in part by the Office of Naval Research [grant numbers N00014-20-1-2141]. The authors would like to thank the State of Ohio to make airborne LiDAR available through Ohio Geographically Referenced Information Program, and Dr.Charles Toth to provide mobile LiDAR data for creating a sub-Figure of Figure 1.

## References

Ahmad Fuad, N., Yusoff, A. R., Ismail, Z., & Majid, Z. (2018). Comparing the performance of point cloud registration methods for landslide monitoring using mobile laser scanning data. *The International Archives of the Photogrammetry, Remote Sensing and Spatial Information Sciences*, *XLII-4/W9*, 11–21.

Aiger, D., Mitra, N. J., & Cohen-Or, D. (2008). 4-points congruent sets for robust pairwise surface registration. *ACM SIGGRAPH 2008 Papers*, 1–10. https://doi.org/10.1145/1399504.1360684

Ao, S., Hu, Q., Yang, B., Markham, A., & Guo, Y. (2021). SpinNet: Learning a General Surface Descriptor for 3D Point Cloud Registration. *Proceedings of the IEEE/CVF Conference on Computer Vision and Pattern Recognition*, 11753–11762.

Arun, K. S., Huang, T. S., & Blostein, S. D. (1987). Least-Squares Fitting of Two 3-D Point Sets. *IEEE Transactions on Pattern Analysis and Machine Intelligence*, *PAMI-9*(5), 698–700. https://doi.org/10.1109/TPAMI.1987.4767965

Bai, X., Luo, Z., Zhou, L., Chen, H., Li, L., Hu, Z., Fu, H., & Tai, C.-L. (2021). PointDSC: Robust Point Cloud Registration using Deep Spatial Consistency. *2021 IEEE/CVF Conference on Computer Vision and Pattern Recognition (CVPR)*, 15854–15864.

Bai, X., Luo, Z., Zhou, L., Fu, H., Quan, L., & Tai, C.-L. (2020). D3Feat: Joint Learning of Dense Detection and Description of 3D Local Features. *Proceedings of the IEEE/CVF Conference on Computer Vision and Pattern Recognition*, 6359–6367.

Barath, D., & Matas, J. (2018). Graph-Cut RANSAC. *Proceedings of the IEEE Conference on Computer Vision and Pattern Recognition*, 6733–6741.
Barrow, H. G., Tenenbaum, J. M., Bolles, R. C., & Wolf, H. C. (1977). *Parametric Correspondence and Chamfer Matching: Two New Techniques for Image Matching*. SRI INTERNATIONAL MENLO PARK CA ARTIFICIAL INTELLIGENCE CENTER.
Besl, P. J., & McKay, N. D. (1992). Method for registration of 3-D shapes. *Sensor Fusion IV: Control Paradigms and Data Structures*, *1611*, 586–606. https://doi.org/10.1117/12.57955
Blekherman, G., Parrilo, P. A., & Thomas, R. R. (Eds.). (2012). *Semidefinite Optimization and Convex Algebraic Geometry*. Society for Industrial and Applied Mathematics. https://doi.org/10.1137/1.9781611972290
Brenner, C., Dold, C., & Ripperda, N. (2008). Coarse orientation of terrestrial laser scans in urban environments. *ISPRS Journal of Photogrammetry and Remote Sensing*, *63*(1), 4–18. https://doi.org/10.1016/j.isprsjprs.2007.05.002
Buch, A. G., Yang, Y., Kruger, N., & Petersen, H. G. (2014). In Search of Inliers: 3D Correspondence by Local and Global Voting. *2014 IEEE Conference on Computer Vision and Pattern Recognition*, 2075–2082. https://doi.org/10.1109/CVPR.2014.266
Bustos, Á. P., & Chin, T.-J. (2018). Guaranteed Outlier Removal for Point Cloud Registration with Correspondences. *IEEE Transactions on Pattern Analysis and Machine Intelligence*, *40*(12), 2868–2882. https://doi.org/10.1109/TPAMI.2017.2773482
Cai, Z., Chin, T.-J., Bustos, A. P., & Schindler, K. (2019). Practical optimal registration of terrestrial LiDAR scan pairs. *ISPRS Journal of Photogrammetry and Remote Sensing*, *147*, 118–131. https://doi.org/10.1016/j.isprsjprs.2018.11.016
Cai, Z., Chin, T.-J., & Koltun, V. (2019). Consensus Maximization Tree Search Revisited. *2019 IEEE/CVF International Conference on Computer Vision (ICCV)*, 1637–1645. https://doi.org/10.1109/ICCV.2019.00172
Campbell, D., Petersson, L., Kneip, L., & Li, H. (2017). Globally-Optimal Inlier Set Maximisation for Simultaneous Camera Pose and Feature Correspondence. *Proceedings of the IEEE International Conference on Computer Vision*, 1–10.
Carlone, L., & Calafiore, G. C. (2018). Convex Relaxations for Pose Graph Optimization With Outliers. *IEEE Robotics and Automation Letters*, *3*(2), 1160–1167. https://doi.org/10.1109/LRA.2018.2793352
Chen, H., & Bhanu, B. (2007). 3D free-form object recognition in range images using local surface patches. *Pattern Recognition Letters*, *28*(10), 1252–1262. https://doi.org/10.1016/j.patrec.2007.02.009
Chen, S., Nan, L., Xia, R., Zhao, J., & Wonka, P. (2020). PLADE: A Plane-Based Descriptor for Point Cloud Registration With Small Overlap. *IEEE Transactions on Geoscience and Remote Sensing*, *58*(4), 2530–2540. https://doi.org/10.1109/TGRS.2019.2952086
Chen, Songlin, Nan, L., Xia, Renbo, Zhao, Jibin, & Wonka, Peter. (2020). PLADE. *IEEE Transactions on Geoscience and Remote Sensing*, *58*(4). https://doi.org/10.1109/tgrs.2019.2952086
Chen, X., & Yu, K. (2019). Feature Line Generation and Regularization From Point Clouds. *IEEE Transactions on Geoscience and Remote Sensing*, *57*(12), 9779–9790. https://doi.org/10.1109/TGRS.2019.2929138
Chen, Y., & Medioni, G. (1992). Object modelling by registration of multiple range images. *Image and Vision Computing*, *10*(3), 145–155. https://doi.org/10.1016/0262-8856(92)90066-C
Cheng, L., Chen, S., Liu, X., Xu, H., Wu, Y., Li, M., & Chen, Y. (2018). Registration of Laser Scanning Point Clouds: A Review. *Sensors*, *18*(5), Article 5. https://doi.org/10.3390/s18051641
Chin, T.-J., Cai, Z., & Neumann, F. (2018). *Robust fitting in computer vision: Easy or hard?* 701–716.
Chin, T.-J., Purkait, P., Eriksson, A., & Suter, D. (2015). Efficient Globally Optimal Consensus Maximisation With Tree Search. *Proceedings of the IEEE Conference on Computer Vision and Pattern Recognition*, 2413–2421.
Choy, C., Dong, W., & Koltun, V. (2020). Deep Global Registration. *Proceedings of the IEEE/CVF Conference on Computer Vision and Pattern Recognition*, 2514–2523.
Chum, O., Matas, J., & Kittler, J. (2003). Locally Optimized RANSAC. *Pattern Recognition*, 236–243. https://doi.org/10.1007/978-3-540-45243-0_31

De Menezes, D. Q. F., Prata, D. M., Secchi, A. R., & Pinto, J. C. (2021). A review on robust M-estimators for regression analysis. *Computers & Chemical Engineering*, *147*, 107254. https://doi.org/10.1016/j.compchemeng.2021.107254

Dempster, A. P., Laird, N. M., & Rubin, D. B. (1977). Maximum Likelihood from Incomplete Data Via the EM Algorithm. *Journal of the Royal Statistical Society: Series B (Methodological)*, *39*(1), 1–22. https://doi.org/10.1111/j.2517-6161.1977.tb01600.x

Deng, Z., Yao, Y., Deng, B., & Zhang, J. (2021). A Robust Loss for Point Cloud Registration. *Proceedings of the IEEE/CVF International Conference on Computer Vision*, 6138–6147.

Deschaud, J.-E., & Goulette, F. (2010). *A Fast and Accurate Plane Detection Algorithm for Large Noisy Point Clouds Using Filtered Normals and Voxel Growing*.

Djahel, R., Vallet, B., & Monasse, P. (2021). Towards Efficient Indoor/Outdoor Registration Using Planar Polygons. *ISPRS Annals of the Photogrammetry, Remote Sensing and Spatial Information Sciences*, *V-2–2021*, 51–58. https://doi.org/10.5194/isprs-annals-V-2-2021-51-2021

Dong, Z., Liang, F., Yang, B., Xu, Y., Zang, Y., Li, J., Wang, Y., Dai, W., Fan, H., Hyyppä, J., & Stilla, U. (2020). Registration of large-scale terrestrial laser scanner point clouds: A review and benchmark. *ISPRS Journal of Photogrammetry and Remote Sensing*, *163*, 327–342. https://doi.org/10.1016/j.isprsjprs.2020.03.013

Eckart, B., Kim, K., & Kautz, J. (2018). HGMR: Hierarchical Gaussian Mixtures for Adaptive 3D Registration. In V. Ferrari, M. Hebert, C. Sminchisescu, & Y. Weiss (Eds.), *Computer Vision – ECCV 2018* (Vol. 11219, pp. 730–746). Springer International Publishing. https://doi.org/10.1007/978-3-030-01267-0_43

Enqvist, O., Ask, E., Kahl, F., & Åström, K. (2012). Robust Fitting for Multiple View Geometry. In A. Fitzgibbon, S. Lazebnik, P. Perona, Y. Sato, & C. Schmid (Eds.), *Computer Vision – ECCV 2012* (pp. 738–751). Springer. https://doi.org/10.1007/978-3-642-33718-5_53

Faugeras, O., & Faugeras, O. A. (1993). *Three-dimensional Computer Vision: A Geometric Viewpoint*. MIT Press.

Fischler, M. A., & Bolles, R. C. (1981). Random sample consensus: A paradigm for model fitting with applications to image analysis and automated cartography. *Communications of the ACM*, *24*(6), 381–395. https://doi.org/10.1145/358669.358692

Fotsing, C., Menadjou, N., & Bobda, C. (2021). Iterative closest point for accurate plane detection in unorganized point clouds. *Automation in Construction*, *125*, 103610. https://doi.org/10.1016/j.autcon.2021.103610

Gao, W., & Tedrake, R. (2019). FilterReg: Robust and Efficient Probabilistic Point-Set Registration Using Gaussian Filter and Twist Parameterization. *2019 IEEE/CVF Conference on Computer Vision and Pattern Recognition (CVPR)*, 11087–11096. https://doi.org/10.1109/CVPR.2019.01135

Ge, X. (2017). Automatic markerless registration of point clouds with semantic-keypoint-based 4-points congruent sets. *ISPRS Journal of Photogrammetry and Remote Sensing*, *130*, 344–357. https://doi.org/10.1016/j.isprsjprs.2017.06.011

Geiger, A., Lenz, P., & Urtasun, R. (2012). Are we ready for autonomous driving? The KITTI vision benchmark suite. *2012 IEEE Conference on Computer Vision and Pattern Recognition*, 3354–3361. https://doi.org/10.1109/CVPR.2012.6248074

Greengard, L., & Strain, J. (1991). The Fast Gauss Transform. *SIAM Journal on Scientific and Statistical Computing*, *12*(1), 79–94. https://doi.org/10.1137/0912004

Guo, Y., Bennamoun, M., Sohel, F., Lu, M., Wan, J., & Kwok, N. M. (2016). A Comprehensive Performance Evaluation of 3D Local Feature Descriptors. *International Journal of Computer Vision*, *116*(1), 66–89. https://doi.org/10.1007/s11263-015-0824-y

Guo, Y., Sohel, F., Bennamoun, M., Lu, M., & Wan, J. (2013). Rotational Projection Statistics for 3D Local Surface Description and Object Recognition. *International Journal of Computer Vision*, *105*(1), 63–86. https://doi.org/10.1007/s11263-013-0627-y

Guru, D. S., Shekar, B. H., & Nagabhushan, P. (2004). A simple and robust line detection algorithm based on small eigenvalue analysis. *Pattern Recognition Letters*, *25*(1), 1–13. https://doi.org/10.1016/j.patrec.2003.08.007

Han, Y., Qin, R., & Huang, X. (2020). Assessment of dense image matchers for digital surface model generation using airborne and spaceborne images – an update. *The Photogrammetric Record*, *35*(169), 58–80. https://doi.org/10.1111/phor.12310
Hartley, R. I., & Kahl, F. (2009). Global Optimization through Rotation Space Search. *International Journal of Computer Vision*, *82*(1), 64–79. https://doi.org/10.1007/s11263-008-0186-9
Hertz, A., Hanocka, R., Giryes, R., & Cohen-Or, D. (2020). PointGMM: A Neural GMM Network for Point Clouds. *2020 IEEE/CVF Conference on Computer Vision and Pattern Recognition (CVPR)*, 12051–12060. https://doi.org/10.1109/CVPR42600.2020.01207
Huang, S., Gojcic, Z., Usvyatsov, M., Wieser, A., & Schindler, K. (2021). *Predator: Registration of 3D Point Clouds With Low Overlap*. 4267–4276. https://openaccess.thecvf.com/content/CVPR2021/html/Huang_Predator_Registration_of_3D_PointClouds_With_Low_Overlap_CVPR_2021_paper.html
Jian, B., & Vemuri, B. C. (2011). Robust Point Set Registration Using Gaussian Mixture Models. *IEEE Transactions on Pattern Analysis and Machine Intelligence*, *33*(8), 1633–1645. https://doi.org/10.1109/TPAMI.2010.223
Ke, Y., & Sukthankar, R. (2004). PCA-SIFT: A more distinctive representation for local image descriptors. *Proceedings of the 2004 IEEE Computer Society Conference on Computer Vision and Pattern Recognition, 2004. CVPR 2004.*, *2*, II–II. https://doi.org/10.1109/CVPR.2004.1315206
Kim, H., & Hilton, A. (2013). Evaluation of 3D Feature Descriptors for Multi-modal Data Registration. *2013 International Conference on 3D Vision - 3DV 2013*, 119–126. https://doi.org/10.1109/3DV.2013.24
Lee, J., Kim, S., Cho, M., & Park, J. (2021). Deep Hough Voting for Robust Global Registration. *Proceedings of the IEEE/CVF International Conference on Computer Vision*, 15994–16003.
Leordeanu, M., & Hebert, M. (2005). *A Spectral Technique for Correspondence Problems Using Pairwise Constraints*. https://doi.org/10.1184/R1/6551327.v1
Li, H. (2009). Consensus set maximization with guaranteed global optimality for robust geometry estimation. *2009 IEEE 12th International Conference on Computer Vision*, 1074–1080. https://doi.org/10.1109/ICCV.2009.5459398
Li, J., Hu, Q., Zhang, Y., & Ai, M. (2022). Robust symmetric iterative closest point. *ISPRS Journal of Photogrammetry and Remote Sensing*, *185*, 219–231. https://doi.org/10.1016/j.isprsjprs.2022.01.019
Li, J., & Lee, G. H. (2019). USIP: Unsupervised Stable Interest Point Detection From 3D Point Clouds. *Proceedings of the IEEE/CVF International Conference on Computer Vision*, 361–370.
Li, J., Zhan, J., Zhou, T., Bento, V. A., & Wang, Q. (2022). Point cloud registration and localization based on voxel plane features. *ISPRS Journal of Photogrammetry and Remote Sensing*, *188*, 363–379. https://doi.org/10.1016/j.isprsjprs.2022.04.017
Li, S., Lu, R., Liu, J., & Guo, L. (2021). Super Edge 4-Points Congruent Sets-Based Point Cloud Global Registration. *Remote Sensing*, *13*(16), Article 16. https://doi.org/10.3390/rs13163210
Lin, Y., Wang, C., Chen, B., Zai, D., & Li, J. (2017). Facet Segmentation-Based Line Segment Extraction for Large-Scale Point Clouds. *IEEE Transactions on Geoscience and Remote Sensing*, *55*(9), 4839–4854. https://doi.org/10.1109/TGRS.2016.2639025
Liu, W., Wu, H., & Chirikjian, G. S. (2021). LSG-CPD: Coherent Point Drift With Local Surface Geometry for Point Cloud Registration. *Proceedings of the IEEE/CVF International Conference on Computer Vision*, 15293–15302.
Lowe, D. G. (2004). Distinctive Image Features from Scale-Invariant Keypoints. *International Journal of Computer Vision*, *60*(2), 91–110. https://doi.org/10.1023/B:VISI.0000029664.99615.94
Mellado, N., Aiger, D., & Mitra, N. J. (2014). Super 4PCS Fast Global Pointcloud Registration via Smart Indexing. *Computer Graphics Forum*, *33*(5), 205–215. https://doi.org/10.1111/cgf.12446
Myronenko, A., & Song, X. (2010). Point Set Registration: Coherent Point Drift. *IEEE Transactions on Pattern Analysis and Machine Intelligence*, *32*(12), 2262–2275. https://doi.org/10.1109/TPAMI.2010.46
Olsson, C., Enqvist, O., & Kahl, F. (2008). A polynomial-time bound for matching and registration with outliers. *2008 IEEE Conference on Computer Vision and Pattern Recognition*, 1–8. https://doi.org/10.1109/CVPR.2008.4587757

Pomerleau, F., Colas, F., Siegwart, R., & Magnenat, S. (2013). Comparing ICP variants on real-world data sets. *Autonomous Robots*, *34*(3), 133–148. https://doi.org/10.1007/s10514-013-9327-2

Qin, Z., Yu, H., Wang, C., Guo, Y., Peng, Y., & Xu, K. (2022). Geometric Transformer for Fast and Robust Point Cloud Registration. *Proceedings of the IEEE/CVF Conference on Computer Vision and Pattern Recognition*, 11143–11152.

Raguram, R., Frahm, J.-M., & Pollefeys, M. (2008). A Comparative Analysis of RANSAC Techniques Leading to Adaptive Real-Time Random Sample Consensus. In D. Forsyth, P. Torr, & A. Zisserman (Eds.), *Computer Vision – ECCV 2008* (pp. 500–513). Springer. https://doi.org/10.1007/978-3-540-88688-4_37

Restrepo, M., & Mundy, J. (2012). An Evaluation of Local Shape Descriptors in Probabilistic Volumetric Scenes. *Procedings of the British Machine Vision Conference 2012*, 46.1-46.11. https://doi.org/10.5244/C.26.46

Rusinkiewicz, S. (2019). A symmetric objective function for ICP. *ACM Transactions on Graphics*, *38*(4), 1–7. https://doi.org/10.1145/3306346.3323037

Rusinkiewicz, S., & Levoy, M. (2001). Efficient variants of the ICP algorithm. *Proceedings Third International Conference on 3-D Digital Imaging and Modeling*, 145–152. https://doi.org/10.1109/IM.2001.924423

Rusu, R. B., Blodow, N., & Beetz, M. (2009). Fast point feature histograms (FPFH) for 3D registration. *Proceedings of the 2009 IEEE International Conference on Robotics and Automation*, 1848–1853.

Schnabel, R., Wahl, R., & Klein, R. (2007). Efficient RANSAC for Point-Cloud Shape Detection. *Computer Graphics Forum*, *26*(2), 214–226. https://doi.org/10.1111/j.1467-8659.2007.01016.x

Segal, A., Haehnel, D., & Sebastian, T. (2009). Generalized-icp. *Robotics: Science and Systems*, *2*(4), 435.

Shi, J., Yang, H., & Carlone, L. (2021). ROBIN: A Graph-Theoretic Approach to Reject Outliers in Robust Estimation using Invariants. *2021 IEEE International Conference on Robotics and Automation (ICRA)*, 13820–13827. https://doi.org/10.1109/ICRA48506.2021.9562007

Sipiran, I., & Bustos, B. (2011). Harris 3D: A robust extension of the Harris operator for interest point detection on 3D meshes. *The Visual Computer*, *27*(11), 963. https://doi.org/10.1007/s00371-011-0610-y

Stamos, I., & Leordeanu, M. (2003). Automated feature-based range registration of urban scenes of large scale. *2003 IEEE Computer Society Conference on Computer Vision and Pattern Recognition, 2003. Proceedings.*, *2*, II–Ii. https://doi.org/10.1109/CVPR.2003.1211516

Tam, G. K. L., Cheng, Z.-Q., Lai, Y.-K., Langbein, F. C., Liu, Y., Marshall, D., Martin, R. R., Sun, X.-F., & Rosin, P. L. (2013). Registration of 3D Point Clouds and Meshes: A Survey from Rigid to Nonrigid. *IEEE Transactions on Visualization and Computer Graphics*, *19*(7), 1199–1217. https://doi.org/10.1109/TVCG.2012.310

Tao, W., Hua, X., Chen, Z., & Tian, P. (2020). Fast and Automatic Registration of Terrestrial Point Clouds Using 2D Line Features. *Remote Sensing*, *12*(8), Article 8. https://doi.org/10.3390/rs12081283

Theiler, P. W., Wegner, J. D., & Schindler, K. (2014). Keypoint-based 4-Points Congruent Sets – Automated marker-less registration of laser scans. *ISPRS Journal of Photogrammetry and Remote Sensing*, *96*, 149–163. https://doi.org/10.1016/j.isprsjprs.2014.06.015

Tombari, F., Salti, S., & Di Stefano, L. (2010). Unique Signatures of Histograms for Local Surface Description. In K. Daniilidis, P. Maragos, & N. Paragios (Eds.), *Computer Vision – ECCV 2010* (pp. 356–369). Springer. https://doi.org/10.1007/978-3-642-15558-1_26

Tombari, F., Salti, S., & Di Stefano, L. (2013). Performance Evaluation of 3D Keypoint Detectors. *International Journal of Computer Vision*, *102*(1), 198–220. https://doi.org/10.1007/s11263-012-0545-4

Tzoumas, V., Antonante, P., & Carlone, L. (2019). Outlier-Robust Spatial Perception: Hardness, General-Purpose Algorithms, and Guarantees. *2019 IEEE/RSJ International Conference on Intelligent Robots and Systems (IROS)*, 5383–5390. https://doi.org/10.1109/IROS40897.2019.8968174

Vizzo, I., Guadagnino, T., Mersch, B., Wiesmann, L., Behley, J., & Stachniss, C. (2022). *KISS-ICP: In Defense of Point-to-Point ICP -- Simple, Accurate, and Robust Registration If Done the Right Way* (arXiv:2209.15397). arXiv. https://doi.org/10.48550/arXiv.2209.15397

Wang, H., Liu, Y., Dong, Z., Wang, W., & Yang, B. (2021). *You Only Hypothesize Once: Point Cloud Registration with Rotation-equivariant Descriptors* (arXiv:2109.00182). arXiv. https://doi.org/10.48550/arXiv.2109.00182

Yang, H., Antonante, P., Tzoumas, V., & Carlone, L. (2020). Graduated Non-Convexity for Robust Spatial Perception: From Non-Minimal Solvers to Global Outlier Rejection. *IEEE Robotics and Automation Letters*, *5*(2), 1127–1134. https://doi.org/10.1109/LRA.2020.2965893

Yang, H., & Carlone, L. (2019). *A Polynomial-time Solution for Robust Registration with Extreme Outlier Rates* (arXiv:1903.08588). arXiv. https://doi.org/10.48550/arXiv.1903.08588

Yang, H., Shi, J., & Carlone, L. (2021). TEASER: Fast and Certifiable Point Cloud Registration. *IEEE Transactions on Robotics*, *37*(2), 314–333. https://doi.org/10.1109/TRO.2020.3033695

Yang, J., Xian, K., Wang, P., & Zhang, Y. (2021). A Performance Evaluation of Correspondence Grouping Methods for 3D Rigid Data Matching. *IEEE Transactions on Pattern Analysis and Machine Intelligence*, *43*(6), 1859–1874. https://doi.org/10.1109/TPAMI.2019.2960234

Yang, J., Xiao, Y., Cao, Z., & Yang, W. (2019). Ranking 3D feature correspondences via consistency voting. *Pattern Recognition Letters*, *117*, 1–8. https://doi.org/10.1016/j.patrec.2018.11.018

Yang, Michael Ying, Förstner, Wolfgang, Department of Earth Observation Science, UT-I-ITC-ACQUAL, & Faculty of Geo-Information Science and Earth Observation. (2010). Plane Detection in Point Cloud Data. In *IGG : Technical Report* (Vols. 1, 2010). University of Bonn. https://research.utwente.nl/en/publications/plane-detection-in-point-cloud-data(dfc1d4a3-c3c4-4da1-867c-ed81d6c32a3b).html

Yuan, W., Eckart, B., Kim, K., Jampani, V., Fox, D., & Kautz, J. (2020). DeepGMR: Learning Latent Gaussian Mixture Models for Registration. In A. Vedaldi, H. Bischof, T. Brox, & J.-M. Frahm (Eds.), *Computer Vision – ECCV 2020* (pp. 733–750). Springer International Publishing. https://doi.org/10.1007/978-3-030-58558-7_43

Zabih, R., & Kolmogorov, V. (2004). Spatially coherent clustering using graph cuts. *Proceedings of the 2004 IEEE Computer Society Conference on Computer Vision and Pattern Recognition, 2004. CVPR 2004.*, *2*, II–II. https://doi.org/10.1109/CVPR.2004.1315196

Zeng, A., Song, S., Niessner, M., Fisher, M., Xiao, J., & Funkhouser, T. (2017). 3DMatch: Learning Local Geometric Descriptors From RGB-D Reconstructions. *Proceedings of the IEEE Conference on Computer Vision and Pattern Recognition*, 1802–1811.

Zhang, J., Yao, Y., & Deng, B. (2021). Fast and Robust Iterative Closest Point. *IEEE Transactions on Pattern Analysis and Machine Intelligence*, 1–1. https://doi.org/10.1109/TPAMI.2021.3054619

Zheng, Y., Sugimoto, S., & Okutomi, M. (2011). Deterministically maximizing feasible subsystem for robust model fitting with unit norm constraint. *CVPR 2011*, 1825–1832. https://doi.org/10.1109/CVPR.2011.5995640

Zhong, Y. (2009). Intrinsic shape signatures: A shape descriptor for 3D object recognition. *2009 IEEE 12th International Conference on Computer Vision Workshops, ICCV Workshops*, 689–696. https://doi.org/10.1109/ICCVW.2009.5457637

Zhou, Q.-Y., Park, J., & Koltun, V. (2016). Fast Global Registration. In B. Leibe, J. Matas, N. Sebe, & M. Welling (Eds.), *Computer Vision – ECCV 2016* (pp. 766–782). Springer International Publishing. https://doi.org/10.1007/978-3-319-46475-6_47